\documentclass[10pt,twocolumn,letterpaper]{article}
\usepackage[utf8]{inputenc}
\usepackage{iccv}
\usepackage{fullpage}
\usepackage{times}
\usepackage{caption}
\usepackage{subcaption}
\usepackage{epsfig}
\usepackage{graphicx}
\usepackage{amsmath}
\usepackage{enumitem}
\usepackage{amssymb}
\usepackage{xcolor}
\usepackage{slashbox}
\usepackage{booktabs}
\newcommand{\omt}[1]{}
\usepackage{algorithmic}
\iccvfinalcopy %
\def\iccvPaperID{24} %

\title{Non-discriminative data or weak model? On the relative importance of data and model resolution}
\author{
    \begin{tabular}{c c c c }
        Mark Sandler &
        Jonathan Baccash &
        Andrey Zhmoginov &
        Andrew Howard \\
        \multicolumn{4}{c}{Google Research} \\            
        \multicolumn{4}{c}{\{sandler, jbaccash, azhmogin, howarda\}@google.com} \\
    \end{tabular}
}
 
\date{\today}
\newcommand{\FIXME}[1]{FIXME{}}
\begin{document}
\maketitle
\begin{abstract}
    We explore the question of how the resolution of the input image (``input resolution'') affects the performance of a neural network when compared to the resolution of the hidden layers (``internal resolution''). Adjusting these characteristics is frequently used as a hyperparameter providing a trade-off between model performance and accuracy. An intuitive interpretation is that the reduced information content in the low-resolution input causes decay in the accuracy.   In this paper, we show that up to a point, the input resolution alone plays little role in the network performance, and it is the internal resolution that is the critical driver of model quality. We then build on these insights to develop novel neural network architectures that we call \emph{Isometric Neural Networks}. These models maintain a fixed internal resolution throughout their entire depth. We demonstrate that they lead to high accuracy models with low activation footprint and parameter count. 
\end{abstract}

\section{Introduction}

Artificial neural networks today are a standard tool for solving many if not most computer vision problems. 
Many different types of neural networks are now running in server farms and mobile devices alike. Once a good architecture is established
a typical design pattern is to {\it scale} such architecture by applying a fixed multiplier to the resolution of each layer, the width of each hidden layer, or the depth (the number of layers) of the network \cite{MobilenetV1,efficientnets}. In the case of resolution multiplier, both the image resolution and the resolution of the inner layers of the architecture are reduced by a given factor, resulting in a faster
but less accurate network. Since lower-resolution images are intuitively less informative than high-resolution ones, the
resulting drop in accuracy is often implicitly attributed to that information loss.  However, the change in the resolution of the hidden layers in the model is an important additional factor that is often ignored.  For example, all of the latest state-of-the-art models 
\cite{efficientnets, fixing-train-test-resolution-discrepancy} use relatively high input resolution, but no acknowledgment of performance with upsampled data. In this paper, we disentangle these two components and demonstrate that the resolution of the input plays a minor role, and it is the internal resolution of the model
that controls model accuracy. We then propose a  novel class of networks that we call {\it Isometric Neural Networks}. The defining characteristic of such models is that they utilize  {\it single}  resolution throughout the entire architecture and consist of {\it identical} blocks stacked on top of each other. We show that these models require fewer parameters and activation memory compared to state of the art models such as MobileNets\cite{MobilenetV1, mobilenetv2, MobilenetV3}.

\section{Related Work}
Image-resolution and model width multipliers have been used extensively both in the literature and by practitioners in the industry to trade computational cost for model accuracy. For instance see \cite{MobilenetV1, mobilenetv2, efficientnets, InceptionV3} among others. In \cite{efficientnets}, an additional layer multiplier was introduced. However, most of this work relies on using high input resolution, which is not always available. In this paper, we disentangle image and model resolutions and show that most of the benefit comes from the internal resolution of the hidden layers. We note that this entanglement has been mentioned before, for example, in section 9 of \cite{InceptionV3}. There, the authors mention the impact of input resolution on InceptionV3 network by adjusting the strides of the earlier layers.  However, it was mentioned as an aside and wasn't explored much further. In this paper, we significantly extend and generalize those results.  To the best of our knowledge, this is the first study focused on evaluating the internal resolution as a factor independent of the input resolution. 

In another related work \cite{fixing-train-test-resolution-discrepancy}, the authors studied the resolution in the context of the relative mismatch
between the test and train resolutions. This direction complements ours. Also related is \cite{downsampled_imagenet}, where it is proposed to
use downsampled ImageNet as an alternative to another low-resolution dataset CIFAR-100 \cite{cifar-100}.
The authors did an early study of the accuracy of low-resolution ImageNet, but they only relied on the width multiplier and the number of layers as the primary ways to parameterize their models. 

Finally, we use models that at low resolution use large receptive fields that arise from the initial conversion of spatial dimension into channel dimension. A related approach relying on atrous convolutions \cite{DeepLabV1} used extensively for semantic segmentation employs similar transformation but uses batch dimension instead.

\section{Our Contributions}
We show that the actual input image resolution plays a minor role in the predictive quality of modern neural networks. Instead, the internal resolution of intermediate tensors is primarily responsible for the trade-off between accuracy and the number of multiply-adds required by the model. Specifically, we show that a model trained with very low-resolution ImageNet images ($14\times 14$) can still achieve respectable accuracy. We also demonstrate that for a fixed input resolution, model accuracy can be further increased without employing more parameters. We then, through carefully designed experiments, eliminate the size of the receptive field as a potential culprit.
    
Secondly, we show that for standard neural
architectures increasing (decreasing) model resolution is
qualitatively equivalent to performing the following three operations: (a) adding (removing)
a few layers at the bottom, (b) applying width multiplier and (c) removing
(adding) a few layers at the top. Such reduction may explain why changing
image resolution and width multiplier often produces very similar trade-off curves.

Our last contribution is a class of novel neural architectures that we call {\it Isometric convolutional networks}.
Isometric networks consist of multiple identical blocks that keep resolution the same throughout the model. An input image of an arbitrary resolution is fed into the first network layer by using the space-to-depth \cite{Shi_2016_CVPR, sajjadi2018frame, yang2019deeperlab} operator or image rescaling. Isometric architectures have multiple appealing properties. First is their simplicity. In particular, they eliminate all pooling layers while still keeping high receptive field. Secondly, isometric networks retain high accuracy while requiring very low inference memory ($<20$\% of MobileNet for the same accuracy, see figure \ref{fig:trade_off_between_accuracy_and_metrics}). Finally, we hope that the simplicity of the isometric architectures might spark more theoretical insights into the model expressiveness in the future.

\section{Resolution and Width Multiplier}
We start by providing a quick review of resolution and width multipliers as standard techniques used for creating a family of models with various performance trade-offs. 

The crux of these techniques is as follows. If we want to get a smaller model, we can reduce the width of each layer (i.e., the number of channels) or the resolution of each layer by a constant factor. Such transformations produce a family of architectures that allow trading accuracy for performance. We show the impact of these changes on model size, activation memory footprint, and MAdds count in table~\ref{table:trade-offs}.  To the best of our knowledge, \cite{MobilenetV1} was the first to articulate these techniques, and they have been extensively used since then, see~\cite{ShuffleNet2017, mobilenetv2, efficientnets} among many others. In particular, \cite{efficientnets} extended this scaling technique to the network depth. The advances in architecture search such as \cite{mnasnet,MobilenetV3} showed that architecture search techniques are capable of discovering more efficient models at particular operating points. However, the multiplier technique remains an essential tool in building families of diverse architectures.  

\begin{figure}[t]
    \centering
    \begin{subfigure}[b]{0.23\textwidth}
    \includegraphics[width=\textwidth]{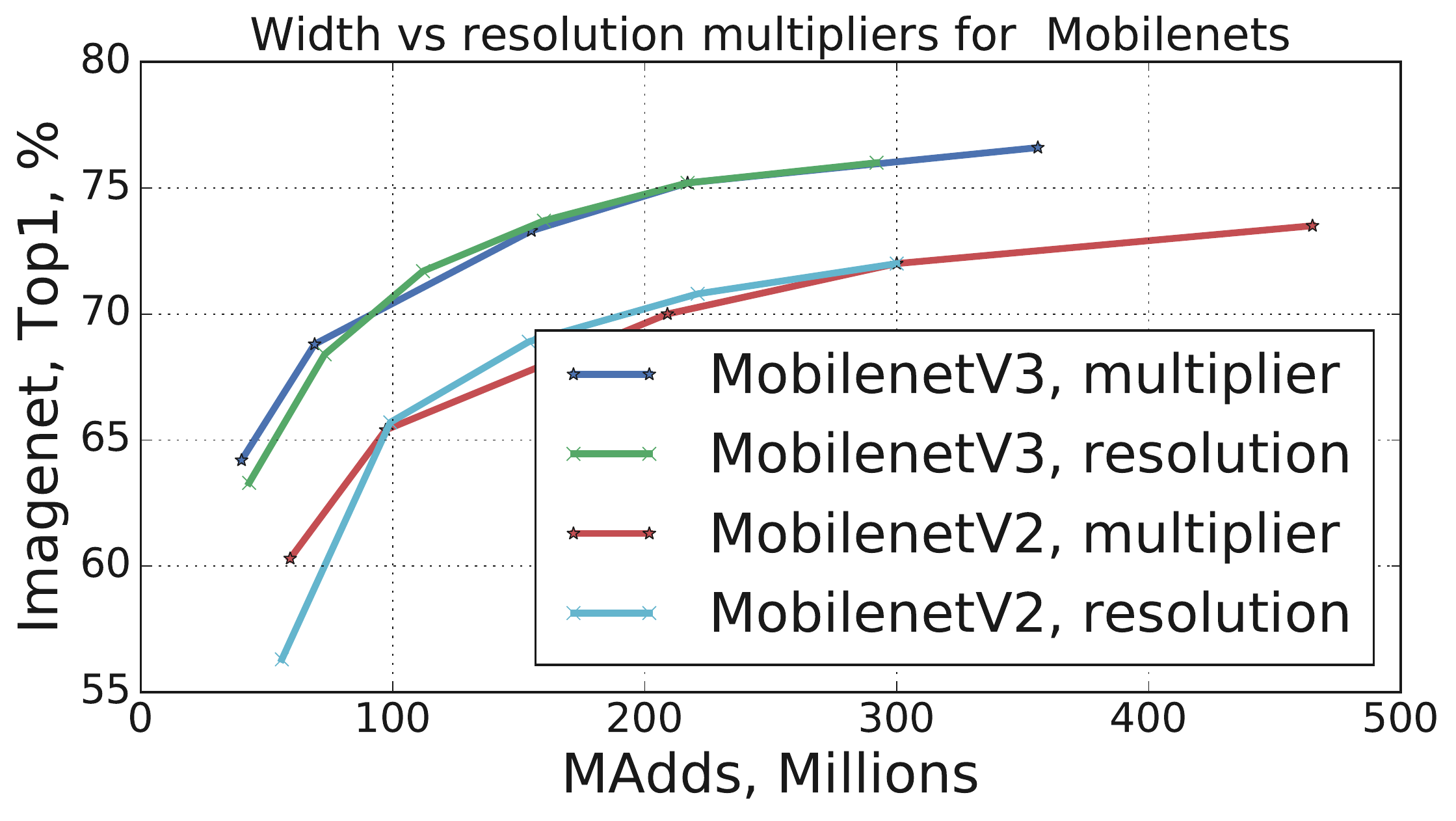}
    \caption{Mobilenets}
    \end{subfigure}
    \begin{subfigure}[b]{0.23\textwidth}
    \includegraphics[width=\textwidth]{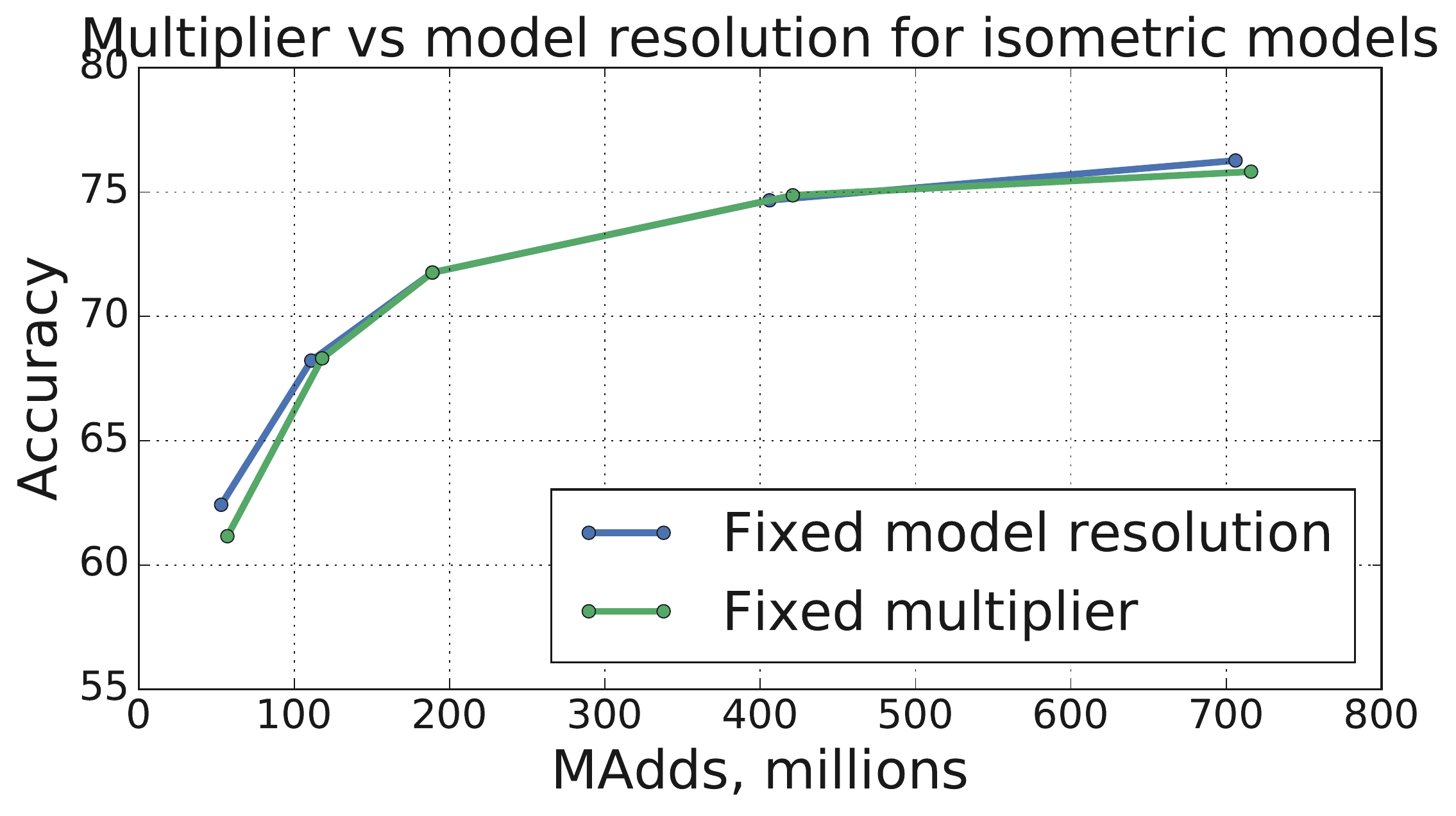}    
    \caption{Isometric}
    \label{fig:isometric_res_vs_width}
    \end{subfigure}
    \caption{Trade-off between width multiplier and image resolution for MobileNets and Isometric networks.}
    \label{fig:mobilenet-multiplier-vs-resolution}
\end{figure}

One interesting empirical property of width and resolution multipliers is that they appear to be nearly fungible when it comes to the trade-off between accuracy and MAdds.\footnote{Throughout the paper we use MAdds -- the number of multiply-and-accumulate in matrix multiplication needed to compute a single inference, as a proxy metric of required computational resources.} That is, changing the resolution by a small factor $\alpha$ has the same impact as changing the width by the same factor, even though one transformation keeps model size constant, while another increases it by a factor $\alpha^2$. For example, for MobileNets we show the trade-off on figure~\ref{fig:mobilenet-multiplier-vs-resolution}.

Here we argue that this empirical property naturally arises from the way the models are built and optimized today. For simplicity consider an architecture $\cal{A}$ whose layers have shape $(r_i \times r_i \times c_i)$, where $r_i$ and $c_i$ are the resolution and number of channels for layer $i$. Most of the commonly used convolutional architectures employ progressive downsampling and increase the number of channels by a factor of 2 
while keeping the number of layers in the same resolution block comparable. 
Therefore we have $r_i = {r_{i-1}}/{s_i}$ and $c_i = s_{i} c_{i-1}$, where
$$
s = {\underbrace{1, \dots 1, 2,}_{b_1 \text{ layers}}  \underbrace{1, \dots 1, 2,}_{b_2\text{ layers}},  \dots} \underbrace{1, \dots 1, 2}_{b_p\text{ layers}}.
$$ 
It then follows that the shape of each tensor in a block $b_p$ is $(r_1/2^{p-1}, 2^{p-1}c_1)$.
Now consider two architectures ${\cal A}_{r=0.5}$ and  
${\cal A}_{c=0.5}$  where we apply either the resolution or the channel width multiplier method. The resulting architectures will then have the following shapes for each layer in a block $b_p$: $(r_1/2^p, 2^{p-1}c_1)$ for ${\cal A}_{r=0.5}$ and $(r_1/2^{p-1}, 2^{p-2}c_1)$ for ${\cal A}_{c=0.5}$. Expanding and aligning these by resolution block  we obtain:
\newcommand{\bl}[1]{b_{#1}\text{ layers} }
\begin{equation}
\begin{array}{ccccc}
 \omt{{\cal A}_{r=0.5}=\{r_i/2, c_i\}_1^k}
 & \underbrace{\{\frac{r_1}{2}, c_i \}}_{\bl{1}} & \dots  & \underbrace{\{\frac{r_1}{2^{p-1}}, 2^{p-2} c_i \}}_{\bl{p-1}} & \underbrace{\{\frac{r_1}{2^{p}}, 2^{p-1} c_i \}}_{\bl{p}}   \\
 \overbrace{\{r_1, \frac{c_i}{2} \}}^{\bl{1}} &  \overbrace{\{\frac{r_1}{2}, c_i \}}^{\bl{2}} & \dots &
 \overbrace{\{\frac{r_1}{2^{p-1}}, 2^{p-2}c_i \}}^{\bl{p}} &  \\
\end{array}
\label{eq:depth-vs-width}
\end{equation}
thus observing that ${\cal A}_{r=0.5}$ is obtained from ${\cal A}_{c=0.5}$ by (a) removing the first block and (b) adding the last block, while also (c) shifting layer counts in each resolution block by 1. For well-calibrated architectures adding or removing a few layers generally results in only minimal changes in performance, and thus it follows that we should expect similar performance when varying width multiplier vs the resolution multiplier.

Finally, we note that despite MAdds and accuracy similarity, width, and resolution multiplier do exhibit very different properties when it comes to model size and activation memory requirements, as shown in table~\ref{table:trade-offs}.
\begin{table}[t]
    \centering
    \begin{tabular}{|l|c|c|c|}
    \hline
         Transformation &  Activation & Model size & MAdds \\
         \hline
        \#channels $\times \alpha$ &  $\alpha$ &$\alpha^2$ & $\alpha^2$ \\
        resolution $\times\alpha$ & $\alpha^2$ & 1 & $\alpha^2$ \\ 
        \#layers $\times \alpha$ & 1 & $\alpha$ & $\alpha$ \\
        \hline
    \end{tabular}
    \caption{Impact of different scaling techniques on model size, activation footprint and MAdds. }
    \label{table:trade-offs}
\end{table}

\section{Resolution Multiplier: Data or Architecture}
\label{sec:lowres-input}
\begin{figure}
\centering       
    \begin{subfigure}[b]{0.23\textwidth}
    \centering
    \includegraphics[clip, trim=1.5cm 1.5cm 3.5cm 3.5cm, width=\textwidth]{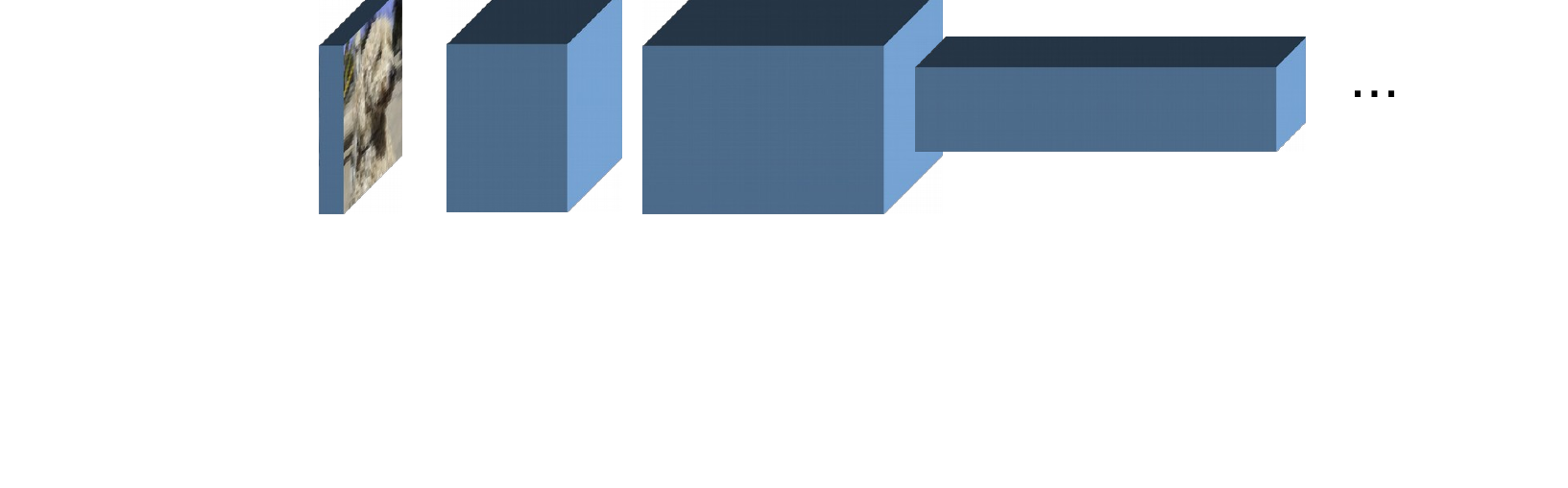}
    \caption{Skip stride}
    \label{skip_stride_adapt}
    \end{subfigure}
    \begin{subfigure}[b]{0.23\textwidth}
    \centering
    \includegraphics[clip, trim=0.5cm 0cm 0.5cm 0.5cm, width=\textwidth]{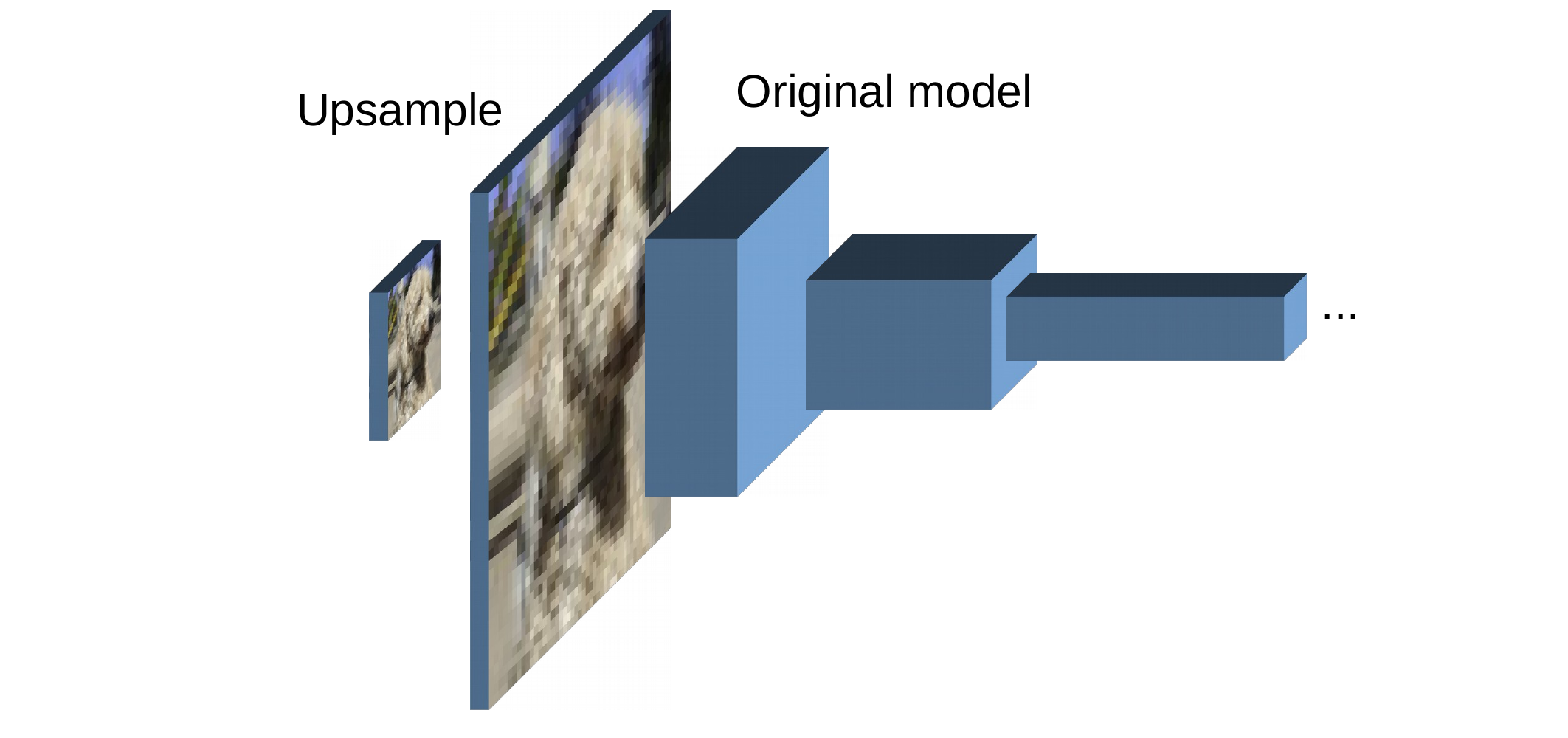}
    \caption{Upsample}
    \label{upsample_adapt}
    \centering
    \end{subfigure}
    \begin{subfigure}[b]{0.23\textwidth}
    \includegraphics[clip, trim=1.5cm 0cm 3.5cm 0cm, width=\textwidth]{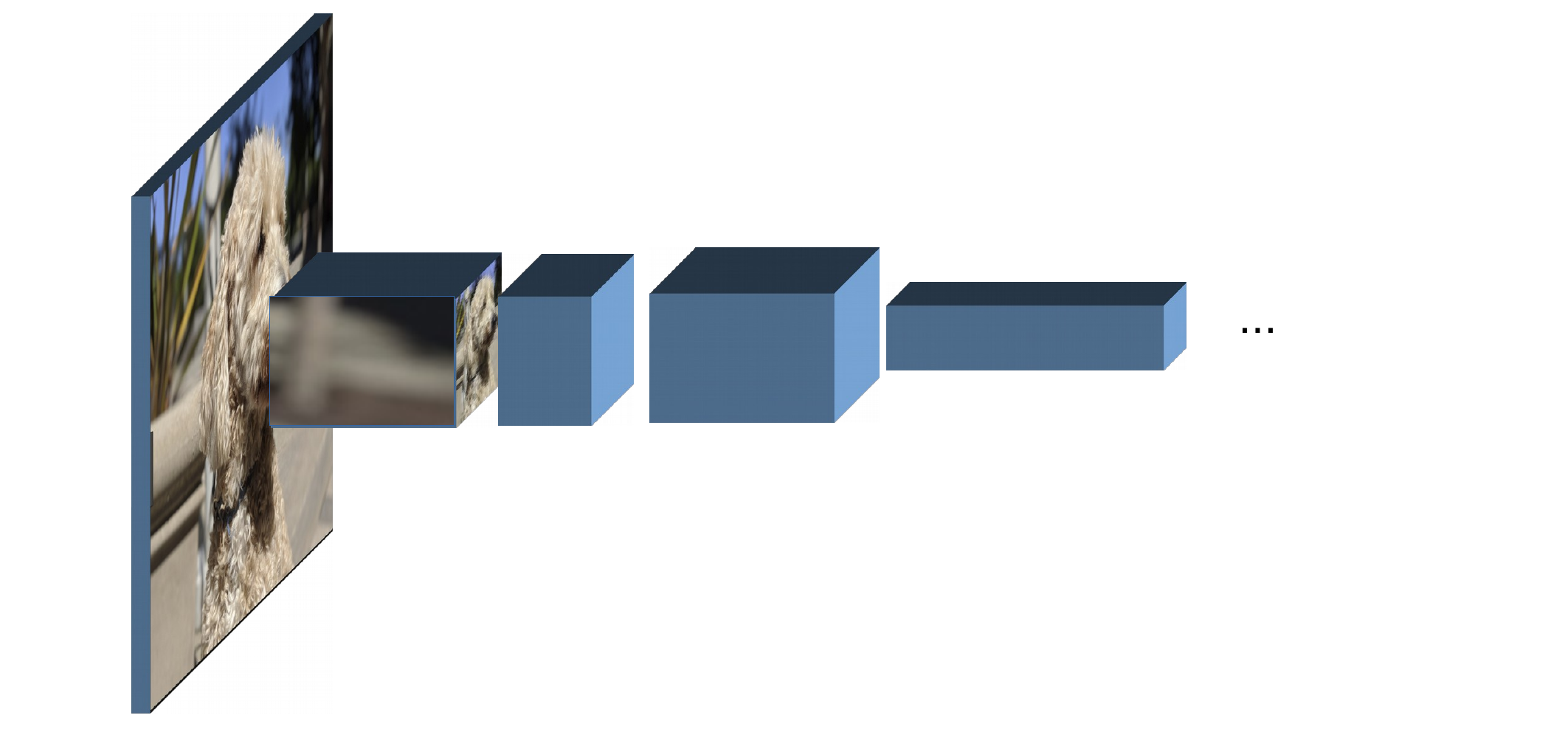}
    \centering
    \caption{S2D + skip stride}
    \end{subfigure}
        \begin{subfigure}[b]{0.23\textwidth}
    \includegraphics[clip, trim=0.5cm 0cm 3.5cm 0cm,width=\textwidth]{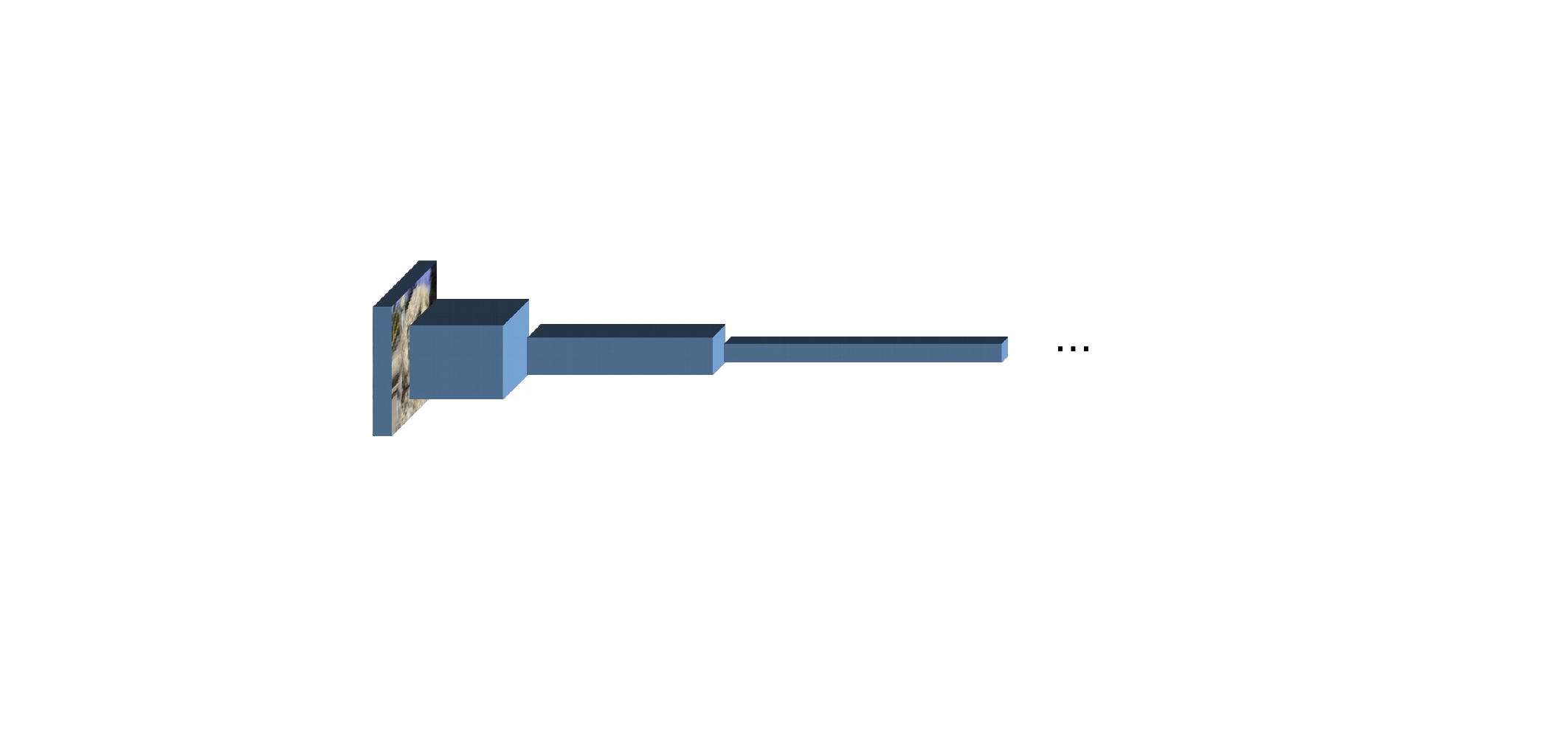}
    \centering
    \caption{Resolution multiplier}
    \end{subfigure}
    \caption{Different ways to feed an image into the model. Here (a) and (b) feed low-resolution input both giving comparable accuracy. Skip-stride (a) eliminates early operators with strides until the resolution matches the original model. Figure (c) shows a variant of skip stride where the input is high resolution but then fed directly into the first low-resolution hidden layer using a space-to-depth transform. In (d) we show standard resolution multiplier for reference.}    
    \label{fig:upsampling-methods}
\end{figure}
\begin{figure}
    \centering
    \begin{subfigure}[b]{0.22\textwidth}
      \centering
      \includegraphics[width=\textwidth]{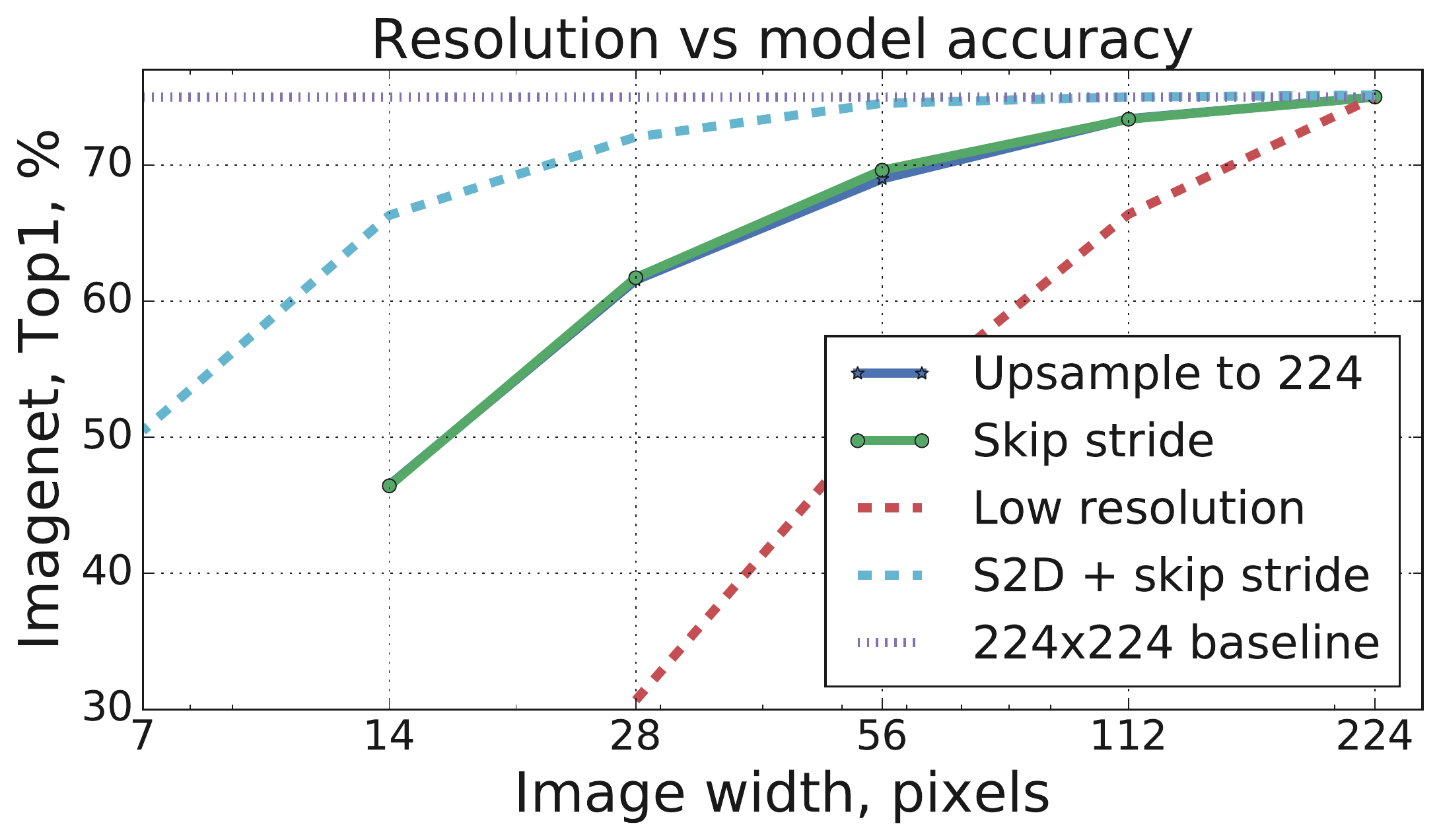}
      \caption{Low resolution}
      \label{lowres}
    \end{subfigure}
    \begin{subfigure}[b]{0.23\textwidth}
      \centering
      \includegraphics[width=\textwidth]{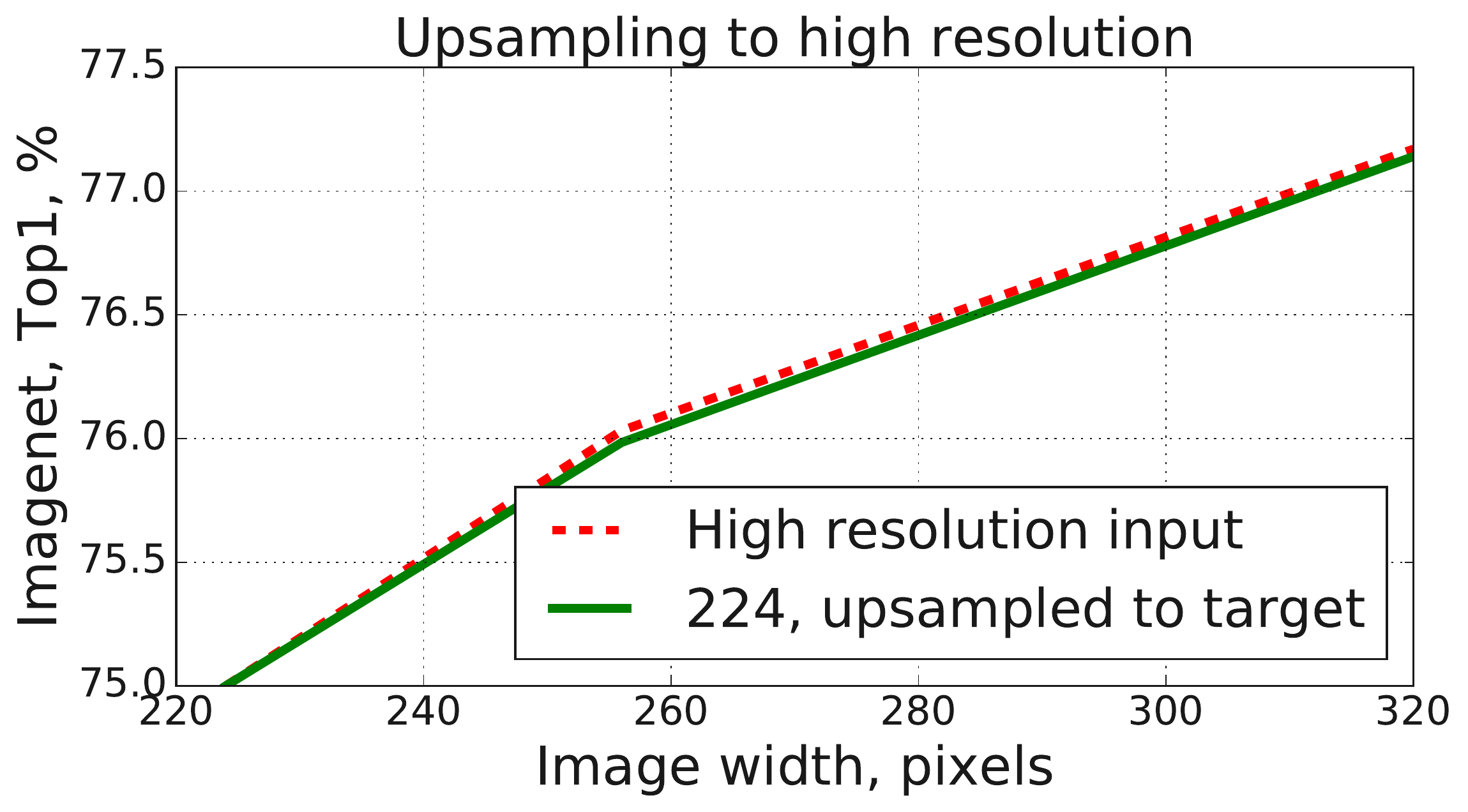}
      \caption{High resolution}
      \label{highres}
    \end{subfigure}
    \caption{The impact of the input size on the model accuracy for MobileNetV3. The low resolution curve in figure \ref{lowres} continues as the high resolution curve in figure \ref{highres}, and corresponds to classical resolution multiplier trade-off. Two nearly matching solid lines in figure \ref{lowres} correspond to bi-linear upsampling and skip-stride methods as described in section \ref{sec:lowres-input}. The cyan dashed line corresponds to using high-resolution input, but with immediate down-sampling using space-to-depth followed by a standard convolution in the first layer. In figure \ref{highres} the green curve corresponds to $224\times 224$ input upsampled to a given resolution. Note that image resolution matters very little compared to the impact of using higher hidden-resolution layers. Best viewed in color.}
    \label{fig:image_resolution}
\end{figure}

In the previous section, we saw that the traditional resolution multiplier essentially acts as a variant of the width-multiplier method with a minor adjustment to layer configuration. Thus it begs the following question: does the input image resolution play a significant role in the model accuracy? Intuitively, the answer seems obvious: a $2\times 2$ image can not be plausibly used to differentiate between different breeds of dogs in ImageNet. On the other hand, $112\times 112$ photo should still retain $99.9$ of the relevant information present in a $399 \times 399$ image. However, despite that, it has been shown multiple times in the literature \cite{efficientnets, InceptionV3, MobilenetV1, mobilenetv2, MobilenetV3} that classifier accuracy grows appreciably as resolution increases. Thus it begs the question: is resolution important, or we are increasing the capacity of the model, and the image resolution was just an artifact of a chosen scaling method? How can we measure these effects?

In this section, we use MobileNetV3\cite{MobilenetV3} as our primary experimental platform. We perform all our experiments on ImageNet\cite{Russakovsky:2015:ILS:2846547.2846559}.

The simplest experiment is to upsample the low-resolution input up to the original resolution. As seen in figure \ref{fig:image_resolution} image upsampling does lead to a dramatic accuracy increase. The
upsampling method itself seems to matter very little, as even dilating the input while filling missing points
with zeros produces the desired result. For instance, $112\times 112$ input resolution leads only to about $1.5\%$ accuracy drop ($73.3\%$ vs. $75.1\%$), compared to nearly $8\%$ drop when the entire model is down-sampled. Similarly,
$56\times 56$ image results in $5\%$ drop when upsampled, and nearly $25\%$ when the model is rescaled. A similar phenomenon has been observed previously, for instance in section 9 of 
InceptionV3~\cite{InceptionV3}, where instead of upsampling the input, the authors skipped initial strides until the image resolution matched the full-resolution model. We illustrate the difference between this and the upsampling in figure \ref{fig:upsampling-methods}. Interestingly, the accuracy of these two methods matched within less than $0.1\%$. It is particularly remarkable that this simple trick allows reaching respectable $45\%$ Top-1, accuracy on image resolution of $14\times 14$, and more than $60\%$ accuracy on image
resolution of $28\times 28$. We assume that by using more powerful models, the accuracy could be pushed even further.

Comparing the similarity in performance between the skip-stride method and the upsampling method, we see that the resolution of the first few layers appears to matter very little. 

To explore the importance of the resolution of the first few layers we perform another experiment. We use the full resolution input and replace the first $3\times 3$ convolutional layer, with a stride-$k$ $3k\times 3k$ convolutional layer. This transformation is also equivalent to a combination of a space-to-depth transformation with block size $k$ followed by the original $3 \times 3$ convolution. If the accuracy does not drop, it would suggest that the model indeed retains most of its discriminative power. For example, to get from $224\times 224$ to $56\times 56$ we use the block size $k=4$. Similarly to get to $7\times 7$, we instead use the block size $k=32$. Strikingly, we see from figure \ref{fig:image_resolution} that
this nearly eliminates any performance losses for resolutions down to $56\times 56$ ($74.6\%$ vs $75.1\%$),  
using just a single linear projection layer, and even allows to downsample down to $7\times 7$ (which corresponds to using $96\times 96$ convolution with stride 32 as a first layer) while still achieving the Top-1 accuracy of nearly $50\%$.  
Such a result is particularly striking when we realize that MobileNetV3's first layer has only 16 channels. So when adapted to $7 \times 7$ resolutions it only contains $7\times 7\times 16=784$ elements, after just a single {\em linear} transformation of the raw image.
These experiments suggest that the high resolutions in the first few layers have only limited utility. On the other hand, the following layers benefit from having a higher internal resolution, as shown by numerous studies on higher-resolution images.
A natural extension of this is to consider an architecture that uses the same
model resolution in all layers. We discuss this idea in the next section.

\section{Isometric Architectures}
\label{sec:isometric-networks}
\begin{figure}[t]
\centering
 \begin{subfigure}[b]{0.22\textwidth}
         \centering
    \includegraphics[width=\textwidth]{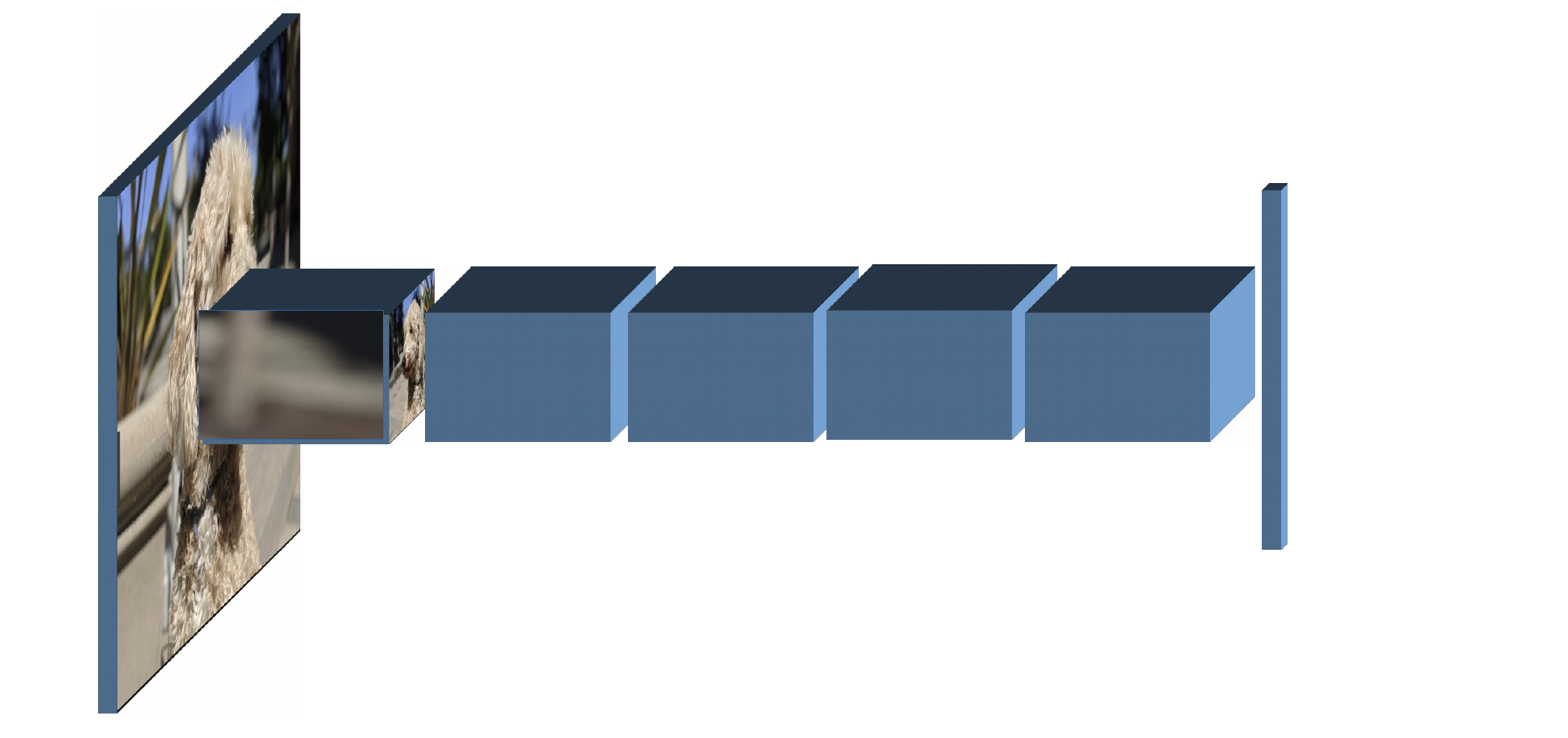} 
\caption{Isometric architecture}
\label{fig:isometric_networks}
\end{subfigure}    
 \begin{subfigure}[b]{0.22\textwidth}
    \includegraphics[width=\textwidth]{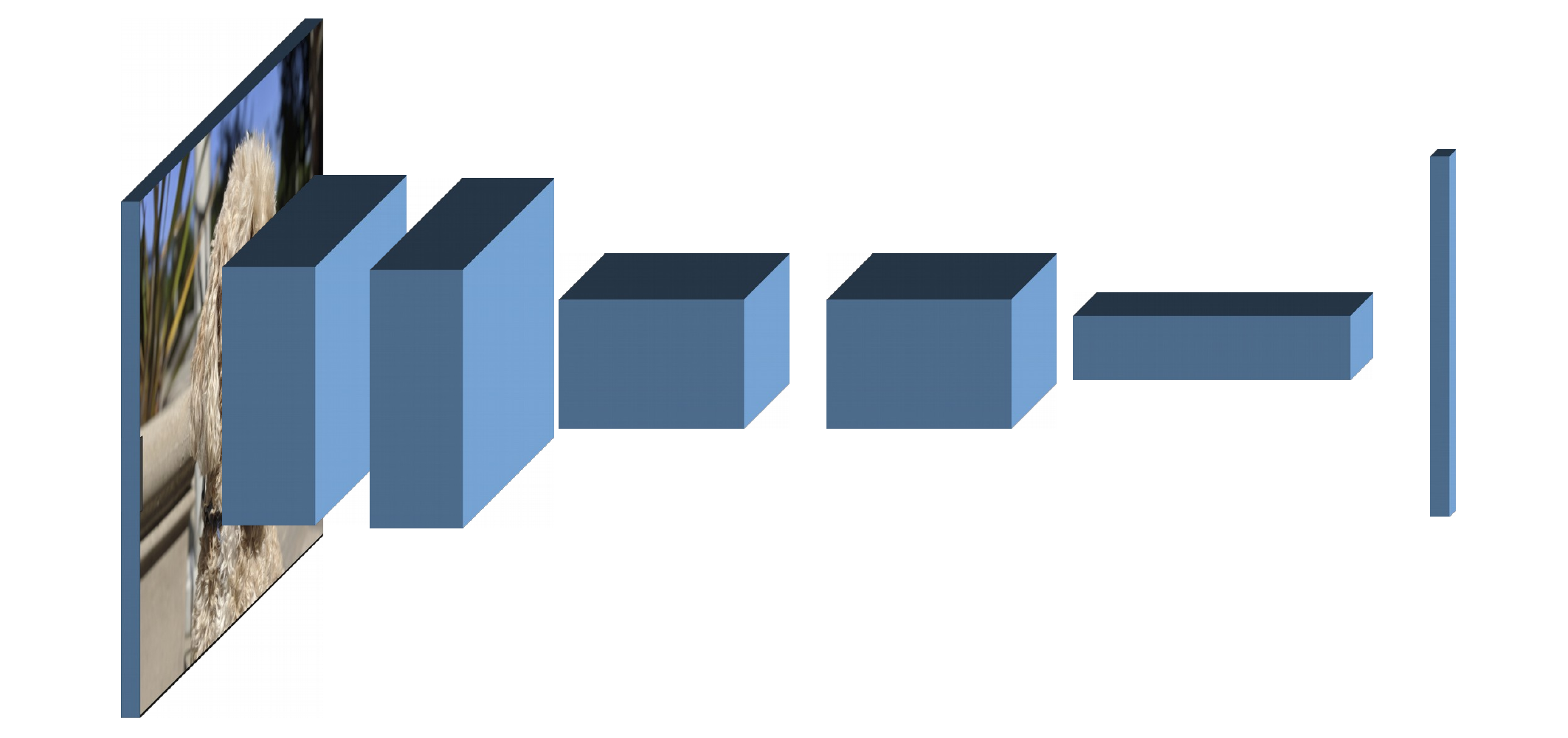}
    \caption{Standard architecture}
\end{subfigure}    
    \caption{Comparison of isometric and standard network architectures. Note that the first layer in isometric architecture could be a down-sampling layer or space-to-depth layer. The impact of the image resolution vs. space-to-depth is shown in Figure \ref{fig:image_resolution}}.
    \label{fig:isometric_vs_classical}
\end{figure}
As discussed in the previous section, using a lower resolution in the first few layers has little impact on model accuracy. Even simply reducing the resolution of all layers down to $7\times 7$ in MobileNetV3 architecture still leads to a somewhat surprising accuracy (50\%), which is higher than what we get if we naively train MobileNet with $56\times 56$ images
(see figure \ref{fig:image_resolution}). We note here that this holds despite introducing a $7\times 7\times 16$ bottleneck in the first layer.  On the other hand, higher-resolution in latter blocks does appear to be helpful as evidenced by the usage of resolution multiplier. What if we built an architecture from scratch that uses only fixed resolution blocks?

In this section, we introduce  {\em isometric architectures}. By design, these architectures maintain constant internal resolution throughout all layers (except for the last global pooling). An illustration of an isometric architecture is shown in figure~\ref{fig:isometric_vs_classical}.

First, let us discuss why such architectures are useful. The primary advantage is that using low resolution 
allows to significantly reduce activation memory footprint, which is essential for many embedded hardware devices. 
Secondly, increasing the internal resolution allows one to build more accurate models while keeping the number of parameters the same. Third, because they use the same block throughout the model, with little cross-layer dependencies, the isometric architectures, can be easily customized to specialized hardware requirements.  For example, 
modern hardware is most efficient when the number of channels is divisible by 32, 64, or even 256. Such a condition is trivial to accommodate in our architectures.  Finally, isometric networks provide a tempting target for theoretical analysis due to their simplicity. In particular, there are no stride operations, and they consist of identical blocks stacked on top of each other in contrast with multi-resolution architectures that employ pooling and striding. At the same time, these models reach nearly 81\% accuracy on ImageNet using standard resolution images, and they are comparable in terms of raw compute requirements to state-of-the-art AutoML models  such as  those in~\cite{efficientnets, MobilenetV3}. 

\begin{figure}
    \centering
    \includegraphics[width=0.4\textwidth]{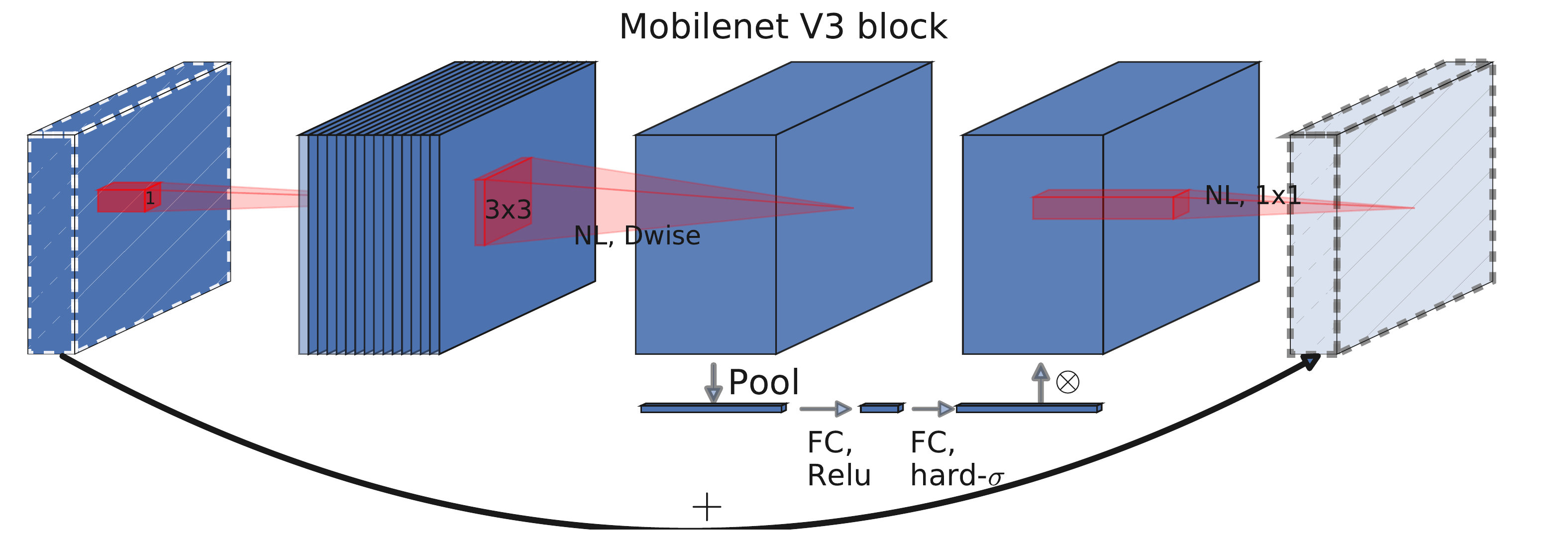}
    \caption{MobileNetV3 \cite{MobilenetV3} block.}
    \label{fig:mobilenet_v3_block}
\end{figure}

In this paper, we begin an empirical evaluation of isomeric networks. We concentrate on the architecture as described in table
\ref{table:isometric-arch}. We note that this only scratches the surface of possible architectures, and also
potentially provides a useful target for NAS-style approaches \cite{NAS_reinforcement, mnasnet} to find better isometric models. 
In our experiments, we use isometric networks consisting of MobileNetV3 + SE bottleneck blocks \cite{MobilenetV3}. For reference, we show the structure of the MobileNetV3 block in figure \ref{fig:mobilenet_v3_block}. We use them mainly as a convenient building block. To match the resolution of the first layer, from
an resolution image that is $k$ time larger, we use the standard space-to-depth operation with the block size $k$. If the first bottleneck uses higher resolution we use bi-linear upsampling instead. Note, space-to-depth transformation, when combined with the first convolutional operation, is equivalent to using a single convolution operator with the kernel size and stride equal to $k$. 

Each block of the isometric network uses a 64 channel bottleneck and uses the expansion factor 6 (that is the expansion size of 384). We conduct experiments with networks containing 8, 16, and 32 identical layers. Since the model is now characterized by a single internal resolution, we will be casually referring to all models that have  $14\times 14$ internal resolution as $14\times 14$ models. 

We note that in the case of isometric neural networks the ``common'' part of the equation \eqref{eq:depth-vs-width} disappears. However, as shown in figure \ref{fig:isometric_res_vs_width}, there is still a remarkable correlation, suggesting that there might be an additional connection between width and image resolution that should be explored.

\begin{table}[t]
    \centering
    \begin{tabular}{l|l}
        Op & Output shape\\ 
        \hline
        Input & $r \times r \times 3$ \\
        \hline
        S2D(target=$d \times d$)& $d\times d \times 3(\frac{r}{d})^2$\\ 
        Conv1x1(64m) & $d\times d \times 64m$ \\ 
      \hline
        $\text{MV3\_SE}(64m, 384m, 4)$ $\times l$ &  $d\times d \times 64m$\\
        Conv1x1($768m$)&   $d\times d \times 768m$ \\
      \hline
        AvgPool(d)      & $1 \times 1 \times 768m$\\
        FC(1280)        & 1280  \\
        \hline
        FC(num\_classes)& num\_classes \\
      \hline
    \end{tabular}
    \caption{Isometric architecture w/MobileNetV3 blocks. We explore architectures containing $l=8$, $16$ and $32$ layers, and utilizing internal resolution $d=7$, $14$ and $28$.  MV3\_SE($x, y,z$) denotes MobileNetV3 block with bottleneck $x$ expansion size $y$ and squeeze-excite factor $z$. Finally, $m$ denotes the 
    width multiplier that we apply to the model backbone. Following
    \cite{mobilenetv2, MobilenetV3} we don't apply multiplier $m$ to fully connected layers.}
    \label{table:isometric-arch}
\end{table}

\subsection{Isometric vs. classical architectures}
\begin{figure}
\centering
 \begin{subfigure}[b]{0.15\textwidth}
         \centering
    \includegraphics[width=\textwidth]{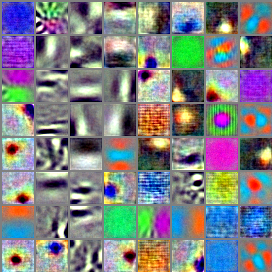} 
    \caption{Type I}
    \label{fig:color-blobs}
\end{subfigure}    
 \begin{subfigure}[b]{0.15\textwidth}
    \includegraphics[width=\textwidth]{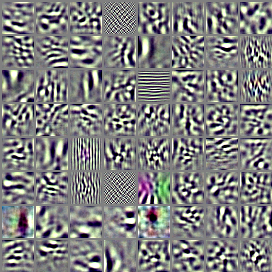}
    \caption{Type II}
\end{subfigure}   
\begin{subfigure}[b]{0.15\textwidth}
\begin{tabular}{|c|c|}
\hline
I  &  62.5\& \\
\hline
II  & 57.2\%  \\
\hline
All &  67.1\% \\
\hline
\end{tabular}
\caption{Accuracy}
\label{fig:firstlayer-accuracy}
\end{subfigure}
\caption{First layer filters for isometric neural network with $32\times 32$ initial
space-to-depth. $1\times 1$ convolution output corresponds to is equivalent to $32\times 32$ convolution applied
to original image. Note how there are two distinct styles of
filters -- the filters which look like very rough edge detectors, and the second that
look like high-resolution maze.  Table (c)  shows the top 1 accuracy
of a model that artificially has frozen features of Type I or Type II respectively.}
\label{fig:firstlayer}
\end{figure}

One intriguing property of isometric networks is that their first layer is essentially a convolution with a huge receptive field. For instance, for $7\times 7$ network, the filter size is 32. Such large filters, present an interesting insight because it is common to assume that the first layer often forms ``edge detectors'' \cite{AlexNet} and other basic feature extractors such as Gabor filters \cite{yosinski2014transferable}. How then would a $32\times 32$ filter look to be general enough to do a large scale recognition?
There appear to be two types of filters. Some look like colorful globs, such as those in figure \ref{fig:color-blobs}, or fairly precise Gabor filters of varying frequency. However, a great number of filters have a complex maze-like regular structure. These filters appear to be a part of an embedding extractor rather than individual feature extractor.\footnote{ The difference is that instead of extracting a single feature, each coordinate in embedding provides a meaningful signal only when looked in combination with other coordinates.}
To measure the relative importance of either type, we trained two architectures with frozen first layers containing respectively colorful blobs and maze-like features. The results are shown in table~\ref{fig:firstlayer-accuracy}. Interestingly, colorful blobs appear to provide
somewhat higher accuracy, but both types of filters contribute significantly to the whole network. Interestingly, in the recent work \cite{faster_neural_networks} 
also observe the maze-like structure of the first layers in architectures that try to learn DCT-like transform and use relatively large $8\times 8$ filters.

\paragraph{Receptive Fields of Isometric Architectures}
The comparison of the receptive field sizes for MobileNetV3 and Isometric networks as a function of depth is reported in figure \ref{fig:receptive-fields}. The fascinating feature of isometric architectures is that the receptive field of convolutional filters spans almost the entire image even in the first few layers.

\section{Experiments}
For our experiments in addition to isometric networks, we also use MobileNetV3 \cite{MobilenetV3}. Our training setup follows that of \cite{MobilenetV3} since our main building block is the same. All our experiments are on ImageNet.

\subsection{Input vs. internal resolution}
In this section, we quantify the importance of the input image versus the internal resolution. For MobileNetV3, the difference between the full resolution model vs. an identical model that uses the upsampled low-res image is shown in figure \ref{fig:image_resolution}. It can be seen that reducing the resolution by a factor of two results in only about 1\% degradation. 

\begin{figure}
    \centering
    \begin{subfigure}[b]{0.23\textwidth}
    \includegraphics[width=\textwidth]{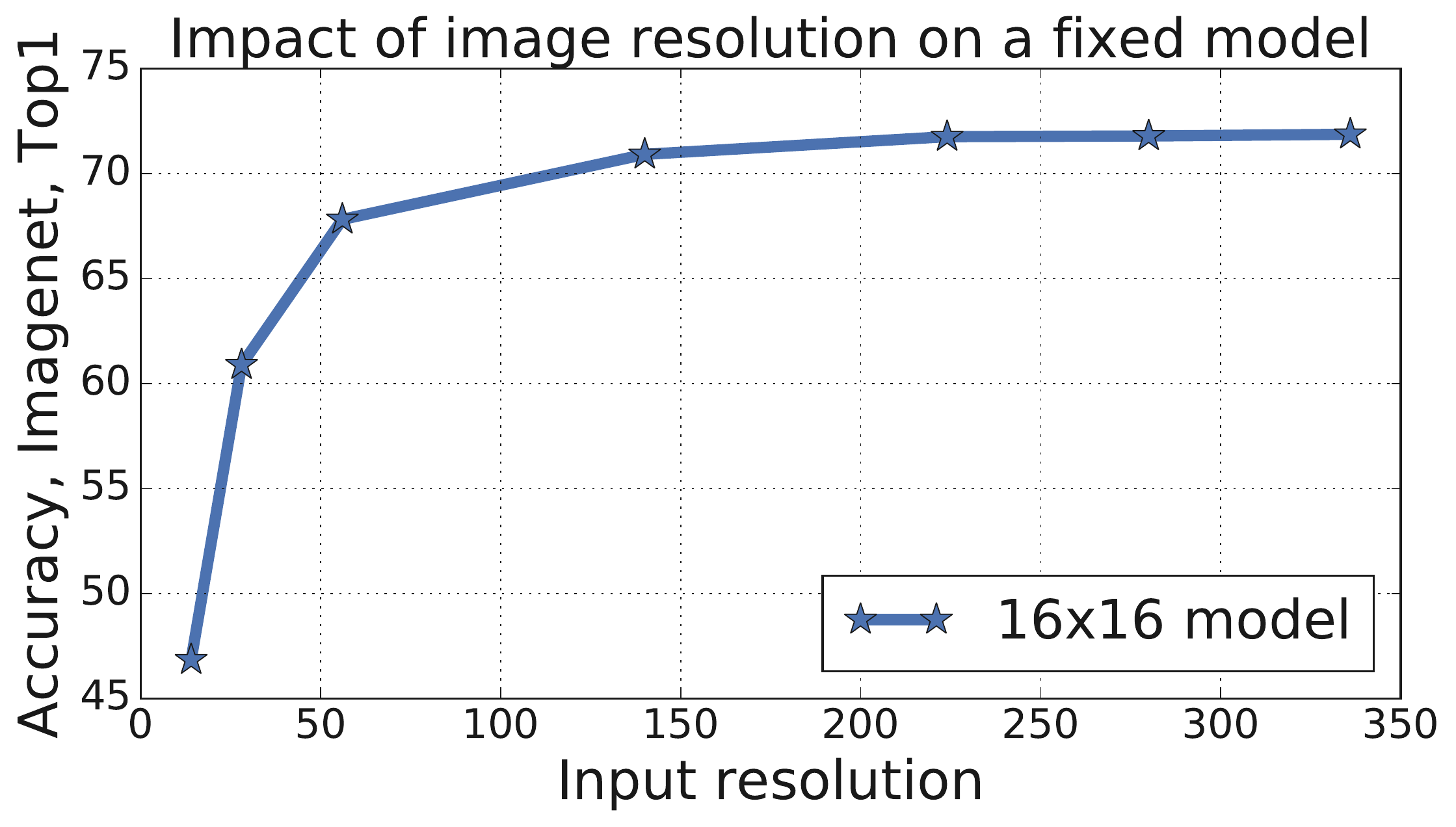}
    \caption{Input resolution}
    \label{fig:fixed_model_change_res}    
    \end{subfigure}
\begin{subfigure}[b]{0.22\textwidth}
\centering
    \includegraphics[width=\textwidth]{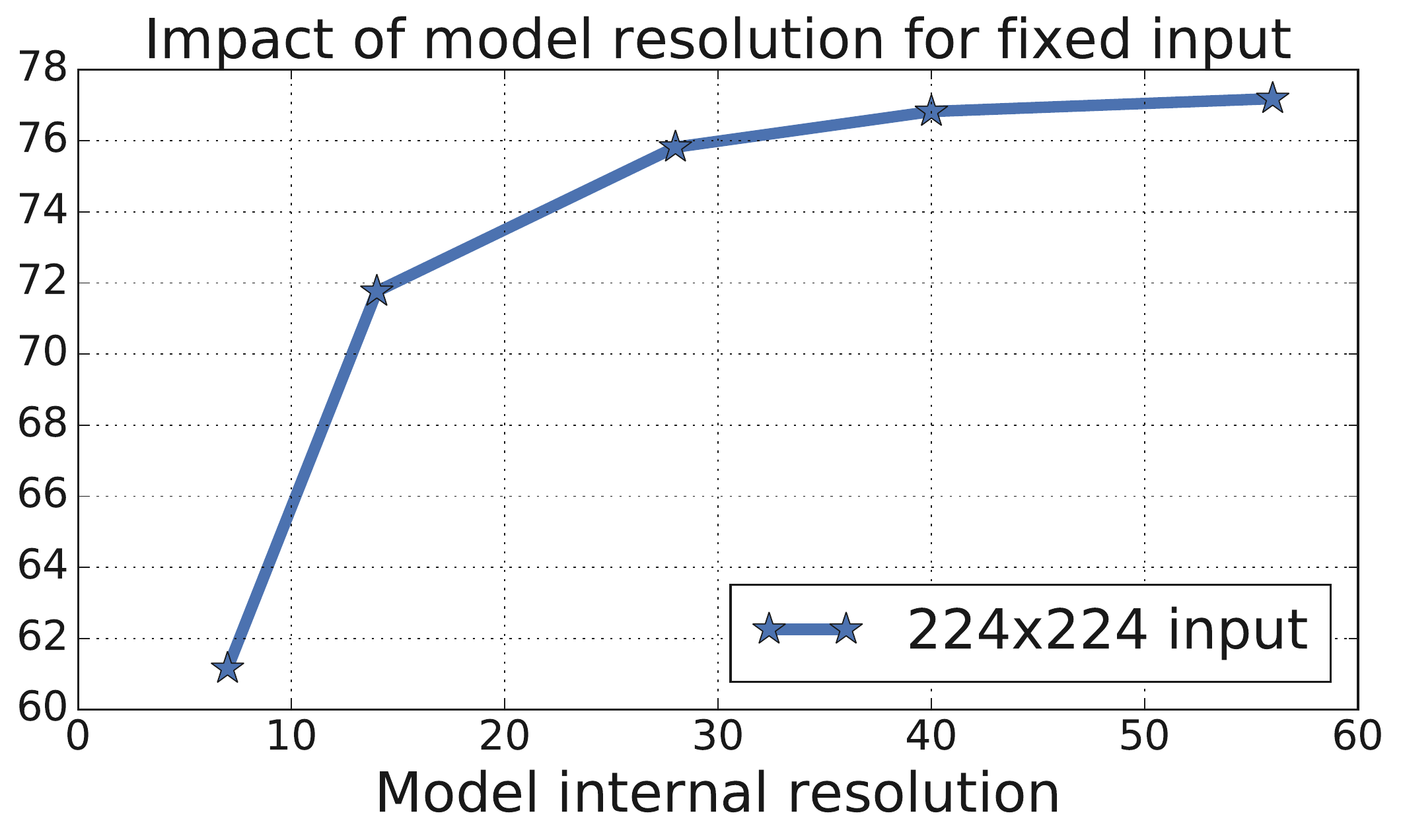}
    \caption{Model resolution}
    \label{fig:fixed_res_change_model}
\end{subfigure}
    \caption{Input vs. model resolution. The baseline $16\times 16$ isometric model with resolution 224 is present on both graphs. Increasing input resolution beyond 224 provides little utility. Increasing model internal resolution improves accuracy significantly, despite not changing the model size or the data. The model has 16 layers and utilizes multiplier 1,
    with $\approx 4.4$ million parameters.}
\end{figure}

Our remaining experiments are with isometric networks. In figure \ref{fig:fixed_res_change_model} we show the trade-off when we
fixed the resolution and instead vary the internal model resolution between
$7\times 7$ and $56\times 56$. In terms of the equivalent resolution of MobileNetV3 that would lead to the same $56\times56$ last layer, 
this corresponds to the input of size  $1792\times 1792$. 

By contrast in figure \ref{fig:fixed_model_change_res} we show the impact 
of resolution on a fixed model, where one can see that there is little benefit of going beyond $224\times 224$ resolution.

Now, we explore the highest accuracy we can achieve with different input
and model resolutions. In table~\ref{tab:internal_vs_input} we compare the
performance of  input resolutions $14\times 14$, $28\times 28$, $56\times
56$ and (the standard) $224\times 224$  against the internal resolutions of
$7\times 7$, $14\times 14$ and $28\times 28$ respectively.  We use an isometric model with 32 layers and multiplier 2.  We use space-to-depth
or upsampling (in case of matching $14\times 14$ input to $28\times 28$
model) to match input to the model. 

Finally, one might ask a question: why change from $14\times 14$ to $28\times 28$ input resolutions produces nearly almost 15\% increase in accuracy, while increase of internal resolution from $14\times14$ to $28\times28$ produces less than 2\% percentage points. Wouldn't it contradict our claim that internal resolution is more important than input? This however should not be surprising, as input resolutions do matter at extreme resolutions. E.g.  one is unlikely to differentiate between 1000 categories using a $4\times4$ sprite.  Internal resolution operate at different scale because hidden layers have much more than 3 channels. For instance $14\times14$ internal resolution corresponds to a large neural network and, with 64 channels,  encodes about as much as $64\times 64\times 3$ image. On the other hand $14\times14$  RGB image corresponds to small input resolution that is rarely if ever used in practice. However once the input is large enough, the model resolution plays a bigger role than the input resolution. 

\subsection{Sizes of receptive fields}
\label{sec:receptive_fields}
\begin{figure}
    \centering
    \begin{subfigure}[t]{0.23\textwidth}
    \includegraphics[width=\textwidth]{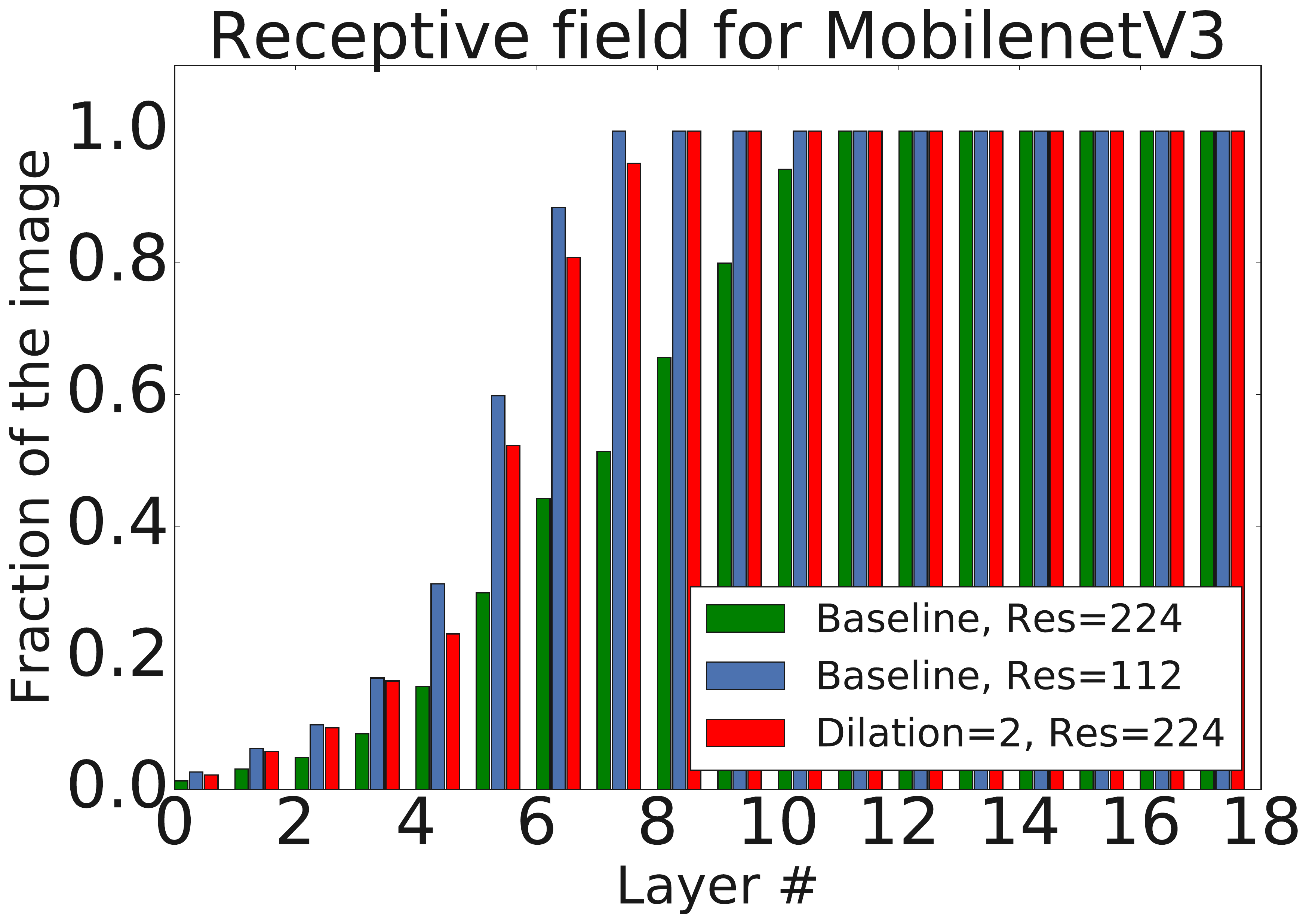}
    \caption{MobileNetV3}
    \end{subfigure}
    \begin{subfigure}[t]{0.23\textwidth}
    \includegraphics[width=\textwidth]{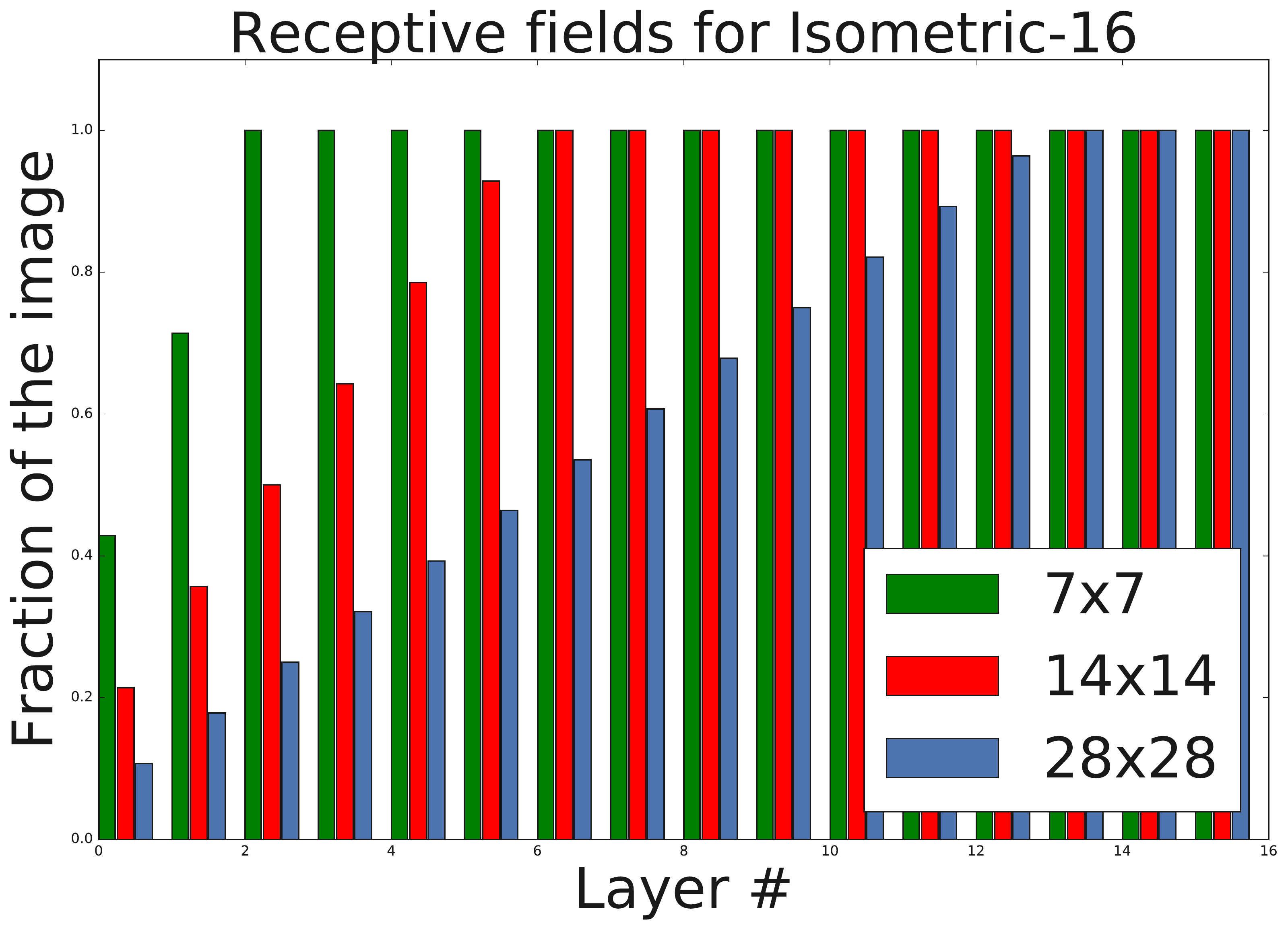}
    \caption{16-layer Isometric}
    \end{subfigure}    
    \caption{The sizes of receptive fields for MobileNetV3 vs Isometric networks. Dilation=2 corresponds 
    to MobileNetV3 with dilated convolution used to mimic the behavior of $112\times 112$ network as described in section \ref{sec:receptive_fields}.}
    \label{fig:receptive-fields}
\end{figure}

Another obvious difference between the high-resolution and low-resolution architectures is the difference in receptive fields. Specifically, low-resolution models have receptive fields that cover a larger fraction of the image compared to high-resolution models. 
It is thus possible that the difference in receptive fields is responsible for the performance difference.
To demonstrate that the receptive field size role is relatively minor, we conduct the following experiment. First, for MobileNetV3, we replace all stride one depthwise convolutional layers with rate two convolutions. This transformation replicates receptive fields of the lower resolution model. Finally  we replace stride=2 layers with $5\times 5$ convolution to maintain changes in receptive fields. The size of the receptive fields is shown in Figure \ref{fig:receptive-fields}.  If the use of the wider receptive field were indeed detrimental, we would observe a significant accuracy drop. However, as shown in table
\ref{tab:dilated-mv3-result}, the actual impact is positive. This experiment strongly suggests that a large receptive field is not a 
source of significant performance degradation, and in fact might have a mild positive impact. 

\begin{table}
    \centering
    \begin{tabular}{|c|c| c|}
    \hline
    dilation 2  & 224 (baseline) & 112(baseline) \\
    \hline
    75.6  & 75.15         & 63  \\
    \hline
    \end{tabular}    
    \caption{Dilated MobileNetV3, Top 1 Accuracy.}
    \label{tab:dilated-mv3-result}
\end{table}

\begin{table}
    \centering
    \begin{tabular}{|c|cccc|}
    \hline
    \backslashbox{\small{MR}}{\small{IR}} &  $14\times 14$ & $28\times 28$ & $56\times 56$ & $224\times 224$ \\
    \hline
$7 \times 7 $ & 48.8 &  62.1 &  67.7 &  70.4 \\
$14 \times 14 $ & 53.7 &  68.2 &  74.7 &  77.6 \\
$28 \times 28 $ & 53.6 &  70.2 &  77.1 &  80.6 \\
    \hline
    \end{tabular}
    \caption{Model vs. input resolution for 32 layer isometric network with multiplier 2 with 20M  parameters. 
    Each row and column corresponds to a different model and input resolution.}
    \label{tab:internal_vs_input}
\end{table}
\begin{figure*}
\centering
\includegraphics[width=0.8\textwidth]{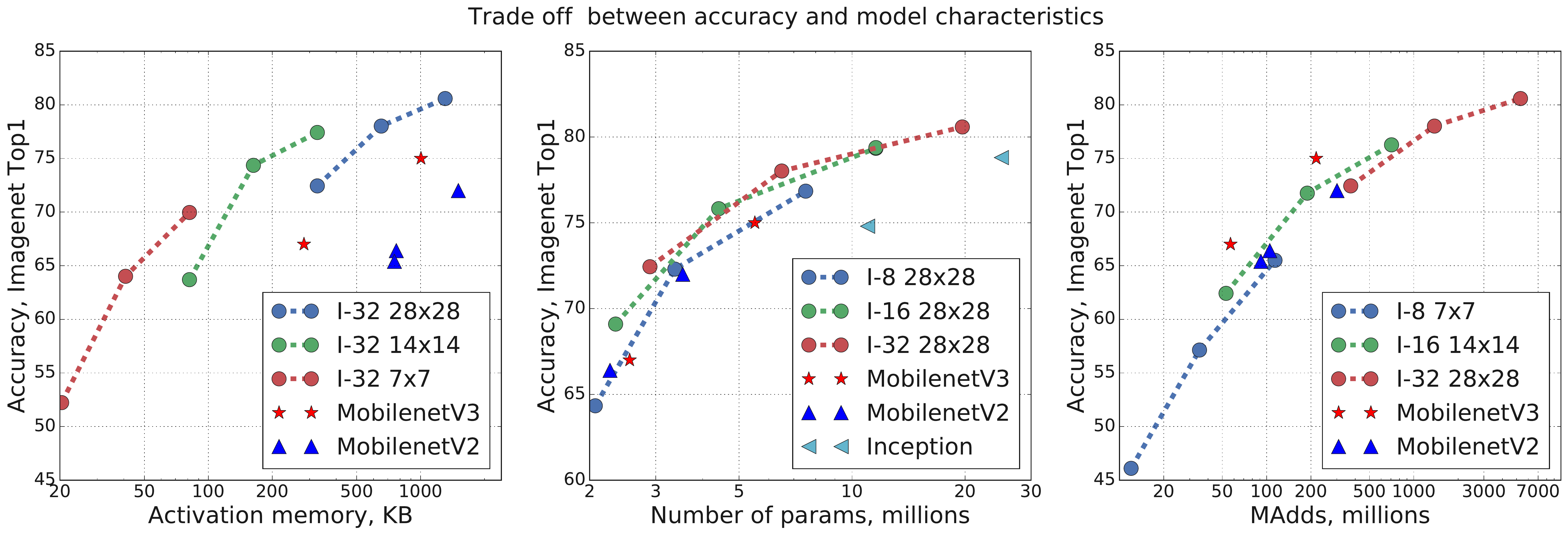}
\caption{Isometric model trade-offs between accuracy and activation memory, model size and MAdds. Each curve contains runs for multipliers 0.5, 1 and 2, for the network with labeled resolution and number of layers. MobileNet numbers are given for reference.}
\label{fig:trade_off_between_accuracy_and_metrics}
\end{figure*}

\subsection{Lowest activation footprint}
The amount of activation memory needed to run inference for isometric models depends on a particular implementation of the inference framework. A good upper bound is a maximum across all layers, the size of all the inputs and outputs.  Thus, to achieve the lowest activation footprint, the obvious strategy is to increase the number of layers, while keeping the internal resolution of the model to a minimum. In figure \ref{fig:trade_off_between_accuracy_and_metrics} we show the performance of the 32-layer models for different multipliers and internal resolutions. The best $7\times 7$ model reaches 70\%+ accuracy while using less than 100K of memory. Similarly, the best $14\times 14$ model
reaches 75\%+ accuracy while using only 300K of memory. By comparison, MobileNetV3 requires nearly 800K to achieve 75\% accuracy.

\subsection{Model size and computation cost}
In figure \ref{fig:trade_off_between_accuracy_and_metrics} we show the trade-off between model size and MAdds and accuracy. For model size, the best
models are those with highest internal resolution. For instance models with 56x56 internal resolution (see figure \ref{fig:fixed_res_change_model}),
achieve 77\% accuracy with less than 5M parameters. For multiply adds we show  comparison in figure~\ref{fig:trade_off_between_accuracy_and_metrics}.
Isometric models, while slightly worse than MobileNetV3, nevertheless provide comparable MAdds vs accuracy trade-off, despite not relying on architecture search.
\begin{figure}
\centering
 \begin{subfigure}[t]{0.15\textwidth}
         \centering
    \includegraphics[width=\textwidth]{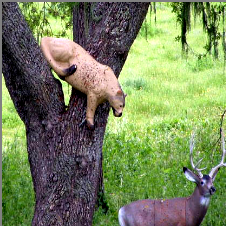}
    \caption{High res input}
\end{subfigure}   
\begin{subfigure}[t]{0.15\textwidth}
    \includegraphics[width=\textwidth]{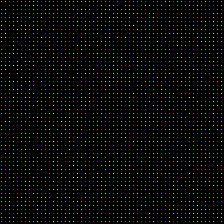}
    \caption{s2b$\rightarrow$s2d mask}
\end{subfigure}   
 \begin{subfigure}[t]{0.15\textwidth}
    \includegraphics[width=\textwidth]{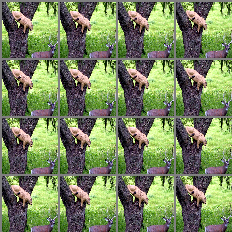}
    \caption{s2b$\rightarrow$s2d batch}
    \label{fig:s2b_s2d_batch}
  \end{subfigure}  
 \begin{subfigure}[t]{0.15\textwidth}
    \centering
    \includegraphics[width=\textwidth]{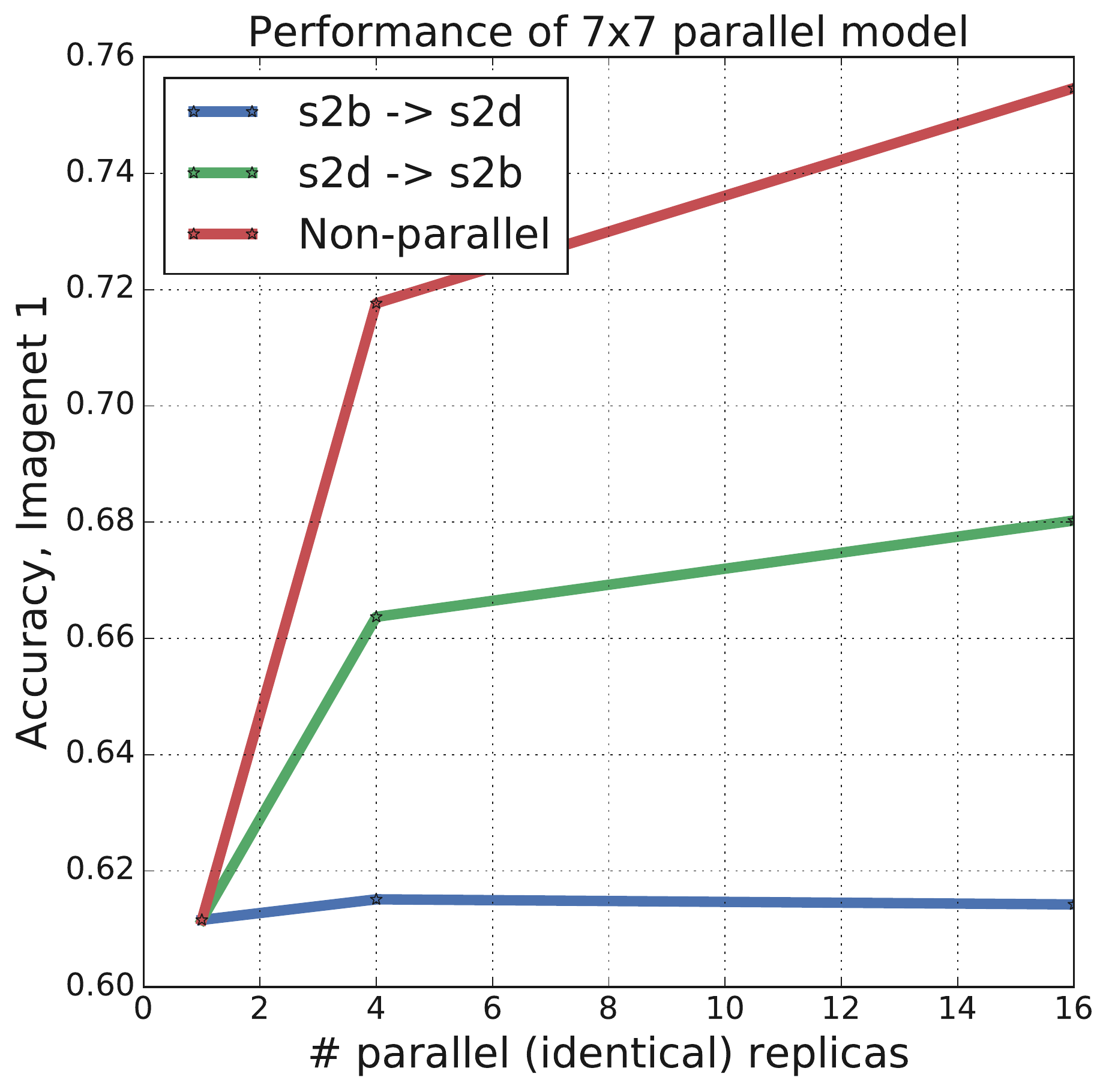}
    \caption{Performance}
    \label{fig:parallel_performance}
 \end{subfigure}
    \begin{subfigure}[t]{0.15\textwidth}
    \includegraphics[width=\textwidth]{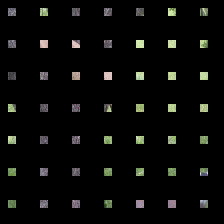}
    \caption{s2d $\rightarrow$ s2b mask}
    \label{fig:s2d_s2b_mask}
  \end{subfigure}    
  \begin{subfigure}[t]{0.15\textwidth}
    \includegraphics[width=\textwidth]{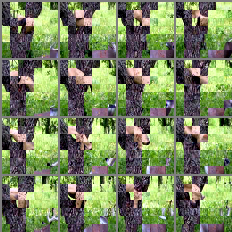}
    \caption{s2d$\rightarrow$s2b batch}
    \label{fig:s2d_s2b_batch}
  \end{subfigure}   
   \caption{The visualization of the input to parallel isometric networks.}
\end{figure}

\subsection{Parallel isometric networks}
From the discussion in section \ref{sec:receptive_fields} we noted that large dilation convolutions actually perform well. However,
part of the reason they even outperform the baseline is the usage of $5\times 5$ convolution in the layers that have stride two. Recall that isometric networks
do not employ stride two layers, thus using dilated convolution with rate $r$ is equivalent to splitting the initial input layer into a batch of ($r^2$) inputs, running them independently in a batch and then averaging out the result. Therefore this converts a single input inference into a large batch size inference, which is often considerably more efficient. Interestingly, a naive split, where we simply split the input into $d\times d$ patches (like those in figure \ref{fig:s2b_s2d_batch}) resulted in only marginal improvement ($\sim 1\%$) as shown in figure \ref{fig:parallel_performance}. Intuitively, it made
sense because the inputs are only marginally different and provided little additional information. However, swapping space-to-batch and space-to-depth, so that we first create a set of patches followed by batching them improved the performance dramatically. This operation is  a shifted application of a grid in figure \ref{fig:s2d_s2b_mask} to generate one sample. The resulting batch of 16 images is shown in figure~\ref{fig:s2d_s2b_batch}. Intuitively this batch forms a data-based ensemble, where the same model is applied to a slightly different view of the data, and the results are averaged out. We show the comparison of these architectures in figure~\ref{fig:parallel_performance}. The baseline curve shows the performance of the model without space-to-batch transformation. We note that such kinds of architectures might be highly beneficial on hardware and frameworks that provide parallel batch compute capability. In that case, using a parallel 16-replica model can boost accuracy, while keeping the wall-time the same. 

\section{Open Questions and Conclusions}
In this paper, we have developed a new way of disentangling neural network internal resolution from the input resolution, and have shown that input resolution plays a fairly minor role in the overall model accuracy. Instead, it is the internal resolution of the hidden layers that are responsible for the impact of
resolution multiplier.  We introduced the notion of {\em isometric convolutional networks} -- the class of neural architectures that share the same resolution throughout the hidden layers. We showed that they are competitive with modern AutoML architectures on several
characteristics, despite their simplicity. Furthermore, since
these architectures lack strides and can potentially share weights across the layers, they form an appealing target for further theoretical analysis.

\section{Acknowledgments}
Authors would like to thank  Michalis Raptis and Alessandro Bissaco for helpful discussions.
{\small
\bibliographystyle{ieee}
\bibliography{bibliography}
}

\end{document}